\documentclass{article}
\usepackage{lineno}
\usepackage[utf8]{inputenc}
\usepackage{amsmath}
\usepackage{caption}
\usepackage{graphicx} 
\usepackage{subcaption}
\usepackage[linesnumbered,ruled,vlined]{algorithm2e}
\usepackage{float}
\usepackage{amsmath,amssymb, amsthm,epsfig,bm}
\usepackage{mathtools}
\usepackage[shortlabels]{enumitem}
\usepackage{caption}
\usepackage{subcaption}
\usepackage[marginal]{footmisc}
\usepackage{authblk}

\title{A Comprehensively Improved Hybrid Algorithm for Learning Bayesian Networks: Multiple Compound Memory Erasing}
\author[1]{Bao-Kui,Mou}
\author[2]{Xin-Hui,Shao\thanks{Corresponding author. Email:{shaoxinhui@mail.neu.edu.cn}.}}
\affil[1]{Department of Mathematics, College of Sciences, Northeastern University, Shenyang, 110819, P.R. China.}
\affil[2]{Department of Mathematics, College of Sciences, Northeastern University, Shenyang, 110819, P.R. China. }
\date{}
\usepackage{amsthm}
\newtheorem{theorem}{Theorem}[section]
\newtheorem{lemma}{Lemma}[section]
\newcommand{\sep}{\!\perp\!\!\!\perp\!}
\newcommand{\sepp}{\perp}
\newcommand{\dsep}[3]{(#1 \sep #2 \!\mid\! #3)_{\mathcal{G}}}
\newcommand{\dsepp}[3]{(#1 \sepp #2 \!\mid\! #3)_{P}}

\newcommand{\indt}[3]{(#1 \sep #2 \!\mid\!#3)}
\usepackage{geometry}
\begin{document}
\maketitle

\begin{abstract}
\label{abs}
Using a Bayesian network to analyze the causal relationship between nodes is a hot spot. The existing network learning algorithms are mainly constraint-based and score-based network generation methods. The constraint-based method is mainly the application of conditional independence (CI) tests, but the inaccuracy of CI tests in the case of high dimensionality and small samples has always been a problem for the constraint-based method. The score-based method uses the scoring function and search strategy to find the optimal candidate network structure, but the search space increases too much with the increase of the number of nodes, and the learning efficiency is very low. This paper presents a new hybrid algorithm, MCME (multiple compound memory erasing). This method retains the advantages of the first two methods, solves the shortcomings of the above CI tests, and makes innovations in the scoring function in the direction discrimination stage. A large number of experiments show that MCME has better or similar performance than some existing algorithms.
\end{abstract}

\par

\section{Introduction}\label{sec1}
\,\,\,\,\,\,\,Bayesian network (BN) is a classical probability graph model. It combines probability theory with graph theory to deal with uncertainty and uses a directed acyclic graph (DAG) to represent the association between nodes. It has been successfully applied to prediction \cite{1}, risk analysis \cite{2}, semantic search, biological system modeling, and other practical fields \cite{3}.This field has two components: the structure learning of Bayesian networks and the parameter learning of Bayesian networks. The latter is based on the former, and the structure learning of Bayesian networks is often more important and complex \cite{4}. BN can enable decision-makers to make conditional causal inferences on uncertain behaviors with the help of the causal relationship of nodes in the network and can deduce the most powerful decision nodes that affect the results. Therefore, this paper focuses on how to generate a prediction network structure with the greatest similarity to the original network structure in a short time, rather than on whether the global score of the prediction model is higher \cite{5}.\par

The methods of learning Bayesian network structure (BNs) from data can generally be divided into three categories: constraint-based, score-based and search strategy, and hybrid algorithms \cite{6}. Representative constraint-based methods mainly include grow-shrink (GS) \cite{7}, three-phase dependency analysis (TPDA) \cite{8}, PC \cite{9}, and incremental associated Markov blanket analysis (IAMB) \cite{10}. The constraint-based methods usually make conditional independence (CI) tests between nodes (i.e. random variables) to find the dependency between nodes, and to determine the skeleton (undirected graph) of the network. For example, the maximum weight spanning tree (MWST) algorithm mentioned in the literature \cite{11} can continuously link the nodes with the maximum mutual information to find an undirected graph without forming a closed-loop and then use the special properties between triples to judge the direction between triples according to mutual information or conditional mutual information. However, with the increase in the number of nodes and the types of nodes, the computational complexity increases exponentially. One disadvantage of constraint-based is that it requires a large number of samples to ensure the effectiveness of CI tests, and the sample size increases too much with the increase of the number of network nodes, which leads to the unreliability of CI tests in the process of learning high-dimensional network structure. What is worse, this early unreliability will lead to correlation effects, making the final learned network structure completely different \cite{12}.\par

The methods of score-based and search strategy mainly use the scoring function to score Bayesian network, to measure the fitness of data and network structure. The scoring functions mainly include K2 \cite{13}, BDe \cite{14}, BIC \cite{15}, and MIT \cite{16}. Combined with the search strategy, the scoring measurement is made for each candidate network structure, and the one with the highest score is selected as the final learned network structure. However, Robinson et al. have proved that the number of network structures to be searched is related to the number of nodes, and when the number of nodes is greater than 5, the amount of calculation for scoring all candidate network structures is very large, and the number of network nodes in actual work will be much greater than 5. To reduce the amount of computation and computation time, Campos et al. used the decomposable property of the BN scoring function to score only the changed local network each time, reducing the time and computation cost, but this did not change the search times \cite{17}. To reduce the number of searching, the ordering-based search (OBS) is widely used. This method is a change based on the scoring method. By sorting the nodes, the parent nodes of each node will only appear in the node-set before the node number, which greatly reduces the network space to be tested at one time. Representative methods include OBS \cite{18}, IINOBS \cite{19}, and WINASOBS \cite{20}.\par

To overcome the shortcomings of the two methods based on constraints and scoring, someone proposed the concept of a hybrid algorithm, hoping to combine the advantages of the two methods through a hybrid algorithm \cite{21}. For example, the MMHC algorithm first uses the CI tests to find the undirected graph from the empty graph and then uses the score-based local search to explore the direction of the edges between nodes, to learn the entire network structure \cite{22}. CB algorithm first uses CI tests to deduce the order of nodes and then uses the modified K2 algorithm to learn the network structure \cite{23}. EGS algorithm first uses PC algorithm to search multiple basic graphs \cite{24}, then randomly converts each basic graph into DAG, uses Bayesian scoring function to score DAG, and finally uses Bayesian scoring function to search the maximum score DAG \cite{25}. IMAPR algorithm also uses CI tests to find a more promising starting point to restart the local structure search \cite{26}.\par

However, the above hybrid algorithms simply combine the two methods and do not solve the inaccurate problem of CI tests in the face of small samples, which is the inherent weakness of applying CI tests in the above hybrid algorithm. This has prompted many scholars to explore new hybrid algorithms, hoping to combine the advantages of the two methods and avoid the shortcomings of the two methods. In this paper, a new hybrid algorithm MCME (Multiple Compound Memory Erasing) is proposed, which combines the advantages of the two methods, overcomes the defect of inaccurate CI tests caused by too many nodes in a small sample, and proposes a new scoring function and scoring method to reduce the time consumption of directional discrimination between nodes.\par

The rest of the paper is organized as follows. The section \ref{sec2} introduces the preparatory knowledge of BN learning. The section \ref{sec3} introduces the defects and improvements of the traditional hybrid algorithm. The section \ref{sec4} introduces the framework of the MCME algorithm. The section \ref{sec5} introduces the experimental evaluation. The section \ref{sec6} concludes the paper and highlights future work.

\section{Preliminary knowledge}
\label{sec2}
\,\,\,\,\,\,\,In this chapter, we will introduce the knowledge of probability theory and information theory related to BN learning. To facilitate the understanding of the subsequent parts of the article, some basic definitions and theorems are re-elaborated.\par
In this paper, we use capital letters (such as $X$ and $Y$) to denote random variables, and lowercase letters (such as $x$ and $y$) to denote the values of random variables. Boldface capital letters (such as $\textbf{X}$ and $\textbf{Y}$) represent sets of random variables, while boldface lowercase letters (such as $\textbf{x}$ and $\textbf{y}$) represent the values of the variable sets. Bayesian network (BN) is a probabilistic graphical model, which can be expressed as $\mathcal{B}=(\mathcal{G},P)$, $\mathcal{G}$ represents the Bayesian network structure (directed acyclic graph DAG), and $P$  represents the conditional probability between all nodes distributed. Where $\mathcal{G}=<\textbf{V},E>$ , $\textbf{V}=(X_1,X_2,\dots,X_n)$ represents the set of nodes in the Bayesian network, and $E$ represents the directed edges between each node. If $P$ satisfies the local directed Markov property \cite{27}: given the set of parent nodes $(P_a(X_i)$ of node $X_i$ in $\mathcal{G}$, $X_i$ is independent of its non-child nodes. Through the Markov condition, the probability distribution $P$ over the set of nodes $\textbf{V}$ of the Bayesian network can be decomposed as follows:
\begin{equation}  
\nonumber
P(X_1,X_2,\dots,X_n)=\prod_{i=1}^{n}P(X_i|P_a(X_i))
\end{equation}

In a DAG, two nodes are said to be adjacent if there is an edge between them linking the two. And a path $l$ in the DAG from $X_i$ to $X_j$ (the two nodes are not directly connected) represents a node sequence with  $X_i$ as the head node and $X_j$ as the tail node. Any two adjacent nodes in the sequence have an edge linking the two in the DAG. If the directions of all the edges on the above path point to $X_j$, then we call that $X_i$ is an ancestor of $X_j$ and $X_j$ is a descendant of $X_i$.\par

\begin{theorem}
\cite{28}\label{tm1}
In $\mathcal{G}=<\textbf{V},E>$, a path $l$ of $X_i$ and $X_j$ is said to be blocked by a set of nodes $\textbf{Z}$ if and only if:
\begin{enumerate}
    \item $l$ contains a chain structure $X_i\to Z\to X_j$ or a fork structure $X_i\gets Z\to X_j$, where $Z\in \textbf{Z}$, or
    \item $l$ contains a collider structure (also called a v-structure) $X_i\to Z\gets X_j$, where $Z\not\in Z$.
\end{enumerate}
\end{theorem}

\begin{theorem}\cite{28}\label{tm2}
If any path of the two nodes $X_i$ and $X_j$ in the DAG is blocked by the node-set $\textbf{Z}$, then $X_i$ and $X_j$ are said to be d-separated by $\textbf{Z}$, denoted as $\dsep{X_i}{X_j}{\textbf{Z}}$, $\textbf{Z}$ is called the d-separator of $X_i$ and $X_j$. $X_i$ and are called d-connected in $\mathcal{G}$ if they are not d-separated by any variable subset of $\textbf{V}$.
\end{theorem}

The concept of d separation can be extended to sets, for example, $\textbf{X}$ and $\textbf{Y}$ are said to be d-separated by $\textbf{Z}$, if any two nodes in sets $\textbf{X}$ and $\textbf{Y}$ are d-separated by set $\textbf{Z}$. The concept of d-separation is different from the concept of conditional independence of two nodes. It is said that two nodes $X_i$ and $X_j$ are conditionally independent under the condition of setting $\textbf{Z}$ over the probability distribution $P$, If $P(X_i,X_j|\textbf{Z})=P(X_i|\textbf{Z})*P(X_j|\textbf{Z})$ , and denoted as $\dsepp{X_i}{X_j}{\textbf{Z}}$. The d-separation indicates the exact link relationship between two nodes in the network graph, which belongs to the definition of graph theory, while conditional independence indicates the independent relationship between two nodes on the basis of probability, which belongs to the definition of probability theory. The two are not equivalent, and the following lemma \ref{la1} and lemma \ref{la2} indicate the connection between them.
\begin{lemma}\cite{29}\label{la1}
For any three disjoint sets of nodes $\textbf{X}$, $\textbf{Y}$, and $\textbf{Z}$ in a DAG, and for all probability distributions $P$, it satisfies the following conditions:
\begin{enumerate}
    \item $\dsep{\textbf{X}}{\textbf{Y}}{\textbf{Z}}$ $\Rightarrow \dsepp{\textbf{X}}{\textbf{Y}}{\textbf{Z}}$ if $\mathcal{G}$ and $P$ are compatible, and
    \item If $\dsepp{\textbf{X}}{\textbf{Y}}{\textbf{Z}}$ holds in all compatible distributions of $\mathcal{G}$, then $\dsep{\textbf{X}}{\textbf{Y}}{\textbf{Z}}$.
\end{enumerate}
\end{lemma}
It can also be seen that the nature of d-separation is stronger than conditional independence. $P$ and $\mathcal{G}$ are said to be faithful to each other if the conditional independence in the probability distribution $P$ is all obtained by applying a local directed Markov condition to $\mathcal{G}$.
\begin{lemma}\cite{30}\label{la2}
For any three disjoint sets of nodes $\textbf{X}$, $\textbf{Y}$ , and $\textbf{Z}$ in a DAG, we say that $\mathcal{B}$ is faithful if $\dsepp{\textbf{X}}{\textbf{Y}}{\textbf{Z}}$ holds if and only if $\dsep{\textbf{X}}{\textbf{Y}}{\textbf{Z}}$  holds.
\end{lemma}
In this paper, it is assumed that the Bayesian networks of interest satisfy the faithfulness condition, and $\indt{\textbf{X}}{\textbf{Y}}{\textbf{Z}}$ is used to represent the d-separation and conditional independence. This assumption ensures the correctness of inferring the true relationship between random variables in DAG by establishing CI tests based on data and is the core assumption of the constraint-based algorithm.

\begin{theorem}\cite{31}\label{tm3}
Assuming that X is a discrete random variable, and $p(x)=Pr(X=x)$ is the probability of the node takes the value of $x$, then the information entropy of the random variable $X$ is defined as:
\begin{equation}  \label{f1}
H(X)=-\sum_{x\in X}p(x)log(p(x))
\end{equation}
Further, the conditional entropy of $X$ under giving the random variable $y$ is:
\begin{equation}  \label{f2}
H(X|Y)=-\sum_{x\in X}\sum_{y\in Y}p(x,y)log(p(x|y))
\end{equation}
\end{theorem}
In the above formula, $p(x,y)$ is the joint probability of random variables $X$ and $Y$, and $p(x|y)$ is the conditional probability of $X$ when $Y$ is known.

\begin{theorem}\cite{32}\label{tm4}
Assuming that there is a pair of discrete random variables $(X,Y)$, the joint probability density function it obeys is denoted as $p(x,y)$, and the marginal probability density functions are respectively $p(x)$ and $p(y)$, the mutual information between two random variables is defined as:
\begin{equation}\label{f3}
I(X,Y)=\sum_{x\in X}\sum_{y\in Y}log(\frac{p(x,y)}{p(x)P(y)}
\end{equation}
Further, given the set of random variable $Z$, the conditional mutual information of $X$ and $Y$ is:
\begin{equation}\label{f4}
I(X,Y|Z)=\sum_{x,y,z}p(x,y,z)log(\frac{p(x,y|z)}{p(x|z)p(y|z)})
\end{equation}
\end{theorem}
In fact, $2N*I(X,Y|Z)$approximately obeys the $\mathcal{X}^2$ distribution \cite{33}. Spirtes, Glymour, and Scheines gave the test statistic $G^2$  of the conditional independence tests \cite{34}. In the detailed formula, the expansion result of $2N*I(X,Y|Z)$ is consistent with the numerical calculation formula of $G^2$\cite{16}, that is:

\begin{equation}
\nonumber
I(X,Y|Z)=\sum_{x,y,z}\frac{N_{xyz}}{N}log(\frac{N_{xyz}N_{z}}{N_{xz}N_{yz}})
\end{equation}
\begin{equation}
\label{f5}
2N*I(X,Y|Z=2\sum_{x,y,z}N_{xyz}log(\frac{N_{xyz}N_{z}}{N_{xz}N_{yz}})=G^2
\end{equation}

Where $N$ is the number of samples, and $N_{xyz}$ is the number of samples under the condition that the value of node $X$ is $x$, $Y$ is $y$, and the value of Condition node-set $Z$ is $z$. Definitions of $N_{xz}$, $N_{yz}$, and $N_{z}$ are equivalent to $N_{xyz}$. It can be seen that $G^2$ is $2N$ times of conditional mutual information in numerical calculation. In view of the above-mentioned relationship between conditional independence and d-type separation, conditional mutual information can also establish an equivalent relationship with d-type separation, which is also an important basis for applying mutual information in some constraint-based algorithms to find whether there is an edge between nodes in a DAG.

\section{Deficiencies and improvements}
\label{sec3}
\,\,\,\,\,\,\,Traditional hybrid algorithms are often divided into two steps: network skeleton (undirected graph) search and identifying the direction of undirected edges. In the process of network skeleton search, the core is to use the CI test to determine whether there is an edge between nodes, and the disadvantage is the failure of the CI tests; (less than 5 nodes) use the score-based method to judge the direction of the edge in the local network, although this solves the problem of a large search space when scoring judgment directly faces all nodes, the computational resources of some complex nodes are Still huge.
\subsection{Failure of CI tests}
\label{sec3.1}
\begin{theorem}\label{tm5}
We denote the minimum association measure between nodes $X$ and $Y$ under a given set $\textbf{Z}$ as $MinAssoc(X,Y|\textbf{Z})$, and the result is equivalent to:
\begin{equation}
\nonumber
MinAssoc(X,Y|\textbf{Z})=\min_{\textbf{S}\in\textbf{Z}}Assoc(X,Y|\textbf{S})
\end{equation}
\end{theorem}
Where $minAssoc(X,Y|\textbf{S})$ is equivalent to the $p$-value of the CI tests for $X$ and $Y$ given $\textbf{S}$, so when $MinAssoc(X,Y|\textbf{S})=0$, the result is equivalent to $(X\sep Y|\textbf{S})$, and then according to Theorem \ref{tm1} and Theorem \ref{tm2}, it is reasonable to think that there is no directly connected edge between $X$ and $Y$, otherwise, there is an edge linking $X$ and $Y$. The $p$-value is rarely 0 in the CI tests, and a significance level $\alpha$ (usually $\alpha$ =0.01) is usually set. When $p\ge \alpha$, the $p$-value can be considered to be approximately equal to 0.In the actual calculation process, the meaning of comparing the test statistic $G^2$ with the upper $\alpha$ quantile of the $\mathcal{X}^{2}$ distribution $\mathcal{X}_{\alpha}^{2}$ ($f_n$) is the same as the above concept, $G^2\le \mathcal{X}_{\alpha}^{2}$ ($f_n$ )  $\equiv p\ge \alpha,f_n=(|D(X)|-1)*(|D(Y)|-1)*\prod_{\textbf{S}\in\textbf{Z}}|D(\textbf{S})|$. $D(X)$ indicates the number of value types of node $X$ \cite{35,36}.\par

The size of $\mathcal{X}_{\alpha}^{2}(f_n)$ is strictly affected by the degree of freedom of the $\mathcal{X}^2$ distribution. With the increase of the conditional node-set $\textbf{Z}$, its growth rate is much faster than the growth rate of the test statistic $G^2$, which is insufficient for the CI tests. The point appears, that even if the nodes $X$ and $Y$ are non-independent and related, with the continuous expansion of the set of conditional nodes (ie nodes), the test results between $X$ and $Y$ will always become independent and irrelevant. Then it will result in an insufficient search for the associated node-set of $X$.\par

\begin{figure}[ht] 
\centering 
\includegraphics[width=0.6\textwidth]{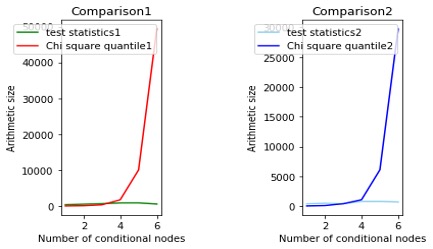} 
\caption{Comparison plot of quantiles and test statistic} 
\label{fig1}
\end{figure}

For example, we apply the PROPERTY dataset \cite{37}, which contains 24 nodes, and we choose the propertyExpenses as the target node $X$, propertyManagement as the node to be tested $Y$, and conditional node-set $\textbf{Z}$ starts from the empty set and continuously adds nodes other than $X$ and $Y$ in $\textbf{V}$ to test whether the CI tests between $X$ and $Y$ will change. The test results are shown in Fig.\ref{fig1}.

The green and light blue line segments in Comparison1 and Comparison2 represent the size of the test statistic, and the red and dark blue line segments represent the corresponding quantile sizes. The nodes contained in the conditional node-set $\textbf{Z}$ in Comparison1 and Comparison2 are different, As a result of more adequate proof.\par

As can be seen from Figure \ref{fig1}, the dependency between nodes $X$ and $Y$ gradually weakens with the increase of the node-set of parents and children. When the size of the set is greater than 2, the dependency between $X$ and $Y$ will disappear. Here, the number of nodes in the node-set of parents and children of $X$ is called the memory capacity of $X$. Through the above experiments, when a set of possible node-set of parents and children of a target node (All nodes with edges existing with $X$ in the undirected graph), and the size of the set exceeds the memory capacity of $X$, even if it encounters a node that has an edge with $X$ in the undirected graph, it cannot be selected correctly. Then, a threshold should be designed for the memory capacity of $X$. When the threshold is exceeded, the memory is cleared for $X$, and then the set of parents and children of $X$ can be found from the remaining nodes. At the same time, the number of memory-erased can be limited to ensure the validity of the searched node-set of parents and children. The details of this improvement are shown in Algorithm \ref{alm1} of section \ref{sec4.2}.\par

\subsection{Insufficiency of mutual information}\label{sec3.2}
\,\,\,\,\,\,\,To describe how much information a random variable contains, the concept of entropy is proposed in information theory. The information entropy of a random variable reflects the amount of information contained in the random variable and is also a measure of the uncertainty of the random variable. Then the joint information entropy of multiple random variables is the total information or uncertainty measure contained in multiple random variables \cite{38}. There is more or less a connection between random variables. To describe the degree of correlation between two or more random variables, the concept of mutual information is proposed in information theory, which describes the public information contained between random variables \cite{39}.

\begin{figure}[ht] 
\centering 
\includegraphics[width=0.4\textwidth]{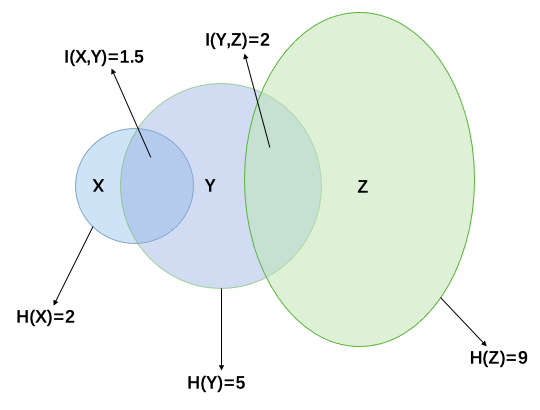} 
\caption{Intersection plot of information between different variables}
\label{fig2}
\end{figure}

Mutual information graph of three nodes, the actual correlation between $X$ and $Y$ is higher than the actual correlation between $Y$ and $Z$.To avoid the shortcomings of CI tests mentioned in \ref{sec3.1}, some scholars choose the node-set of parents and children of the target node by comparing the mutual information between different nodes to be tested and the target node, but it is inaccurate. For example, if $I(Y,Z)>I(X,Y)$ as shown in Fig.\ref{fig2}, then it should be considered that the degree of correlation between $Y$ and $Z$ is higher than that between $X$ and $Y$, but the fact is just the opposite. The fundamental reason is that mutual information is affected by the size of the information entropy of the random variable itself. Since the mutual information satisfies $I(X,Y)\le \min\left\{H(X),H(Y)\right\}$, when the information entropy difference between variables is large, the mutual information is used to judge whether there is a relationship between nodes is inaccurate.\par

\subsection{Entropy-eliminated mutual information}\label{sec3.3}
\,\,\,\,\,\,\,The mutual information between two nodes is affected by the respective information entropy of the two nodes. To make the size comparison between the two mutual information more accurate, the entropy should be eliminated by dividing the mutual information by the information entropy. But for mutual information, corresponds to the information entropy of at least two nodes. For example, we will obtain two ratio values by dividing the mutual information of node $X$ and node $Y$ by the information entropy of the two nodes respectively, denoted as $\eta_{1}=\frac{I(X,Y)}{H(X)}$,$\eta_{2}=\frac{I(X,Y)}{H(Y)}$. $\eta_{1}$ and $\eta_{2}$ can be regarded as the ratio of mutual information in the information entropy of $X$ and $Y$ nodes respectively. The smaller the node information entropy is, the higher the corresponding mutual information ratio is. We define the entropy-eliminated mutual information of two nodes as:

\begin{equation}\label{f6}
EEMI_{XY}=p_{1}\eta_{1}+p_{2}\eta_{2}
\end{equation}

The values of $p_1$ and $p_2$ can be considered to be equal to $\frac{H(Y)}{H(X)+H(Y)}$ and $\frac{H(X)}{H(X)+H(Y)}$, respectively. The assignment method is a parameter self-adjustment mechanism, which can independently assign a larger weight to a larger mutual information ratio, which is of practical significance. The mutual information ratio of a node with smaller information entropy can better reflect the mutual information ratio. The importance of information, and then more able to show the correlation between two nodes, so give a greater weight.\par

We use the AISA data set in the literature \cite{37} for testing. It is known that there are eight variable nodes in the data set, namely: aisa, tub, smoke, lung, bronc, either, xray, and dysp, and smoke has a direct connection with lung and bronc. contact directly. We separately calculate the mutual information and entropy-eliminated mutual information between the node smoke and other nodes.\par

\begin{table}[ht]
\centering
\caption{Comparison of entropy-eliminated mutual information and mutual information}
\setlength{\tabcolsep}{1mm}
\begin{tabular}{cccccccc}
\hline
 & asia & tub & lung & bronc & either & xray & dysp\\
 \hline
 MI & $3.75\times10^{-5}$ & $6.82\times10^{-4}$ & $3.14\times10^{-2}$ & $8.52\times10^{2}$ & $2.47\times10^{-2}$ & $1.43\times10^{-2}$ & $4.20\times10^{-2}$ \\
 EEMI & $5.04\times10^{-4}$ & $8.61\times10^{-3}$ & $8.01\times10^{-2}$ & $8.53\times10^{-2}$ & $5.74\times10^{-2}$ & $2.38\times10^{-2}$ & $4.23\times10^{-2}$\\
 \hline
\end{tabular}
\label{t1}
\end{table}

In table \ref{t1}, MI stands for mutual information, and EEMI stands for entropy-eliminated mutual information. The larger the value, the higher the degree of association with the target node. As can be seen from table \ref{t1}, if the mutual information is directly used to judge, the correct adjacent node bronc can be selected in the first screening, but the wrong node dysp will be selected in the second screening. If we use the entropy-eliminated mutual information proposed above to filter adjacent nodes, two correct adjacent nodes, bronc, and lung, can be selected after two screenings. The above experiments can illustrate the rationality of our improvement of mutual information.\par

In the network skeleton search process of the MCME algorithm, not only the CI tests with memory in section \ref{sec3.1} is considered, but also the entropy-eliminated mutual information is added as a complementary screening process. Experiments show that this supplementary addition is reasonable.\par

\subsection{Orientation discrimination of undirected edges}\label{sec3.4}
\,\,\,\,\,\,\,In the traditional hybrid algorithm, the direction of the edges between the nodes in the network skeleton is a local scoring method. By decomposing the undirected graph (network skeleton) into subgraphs, then applying a score-based approach to search for directed edges between nodes on the subgraphs, and then merging the directed subgraphs to form a complete DAG. This avoids the problem of too large a search space caused by too many network nodes, but it does not fundamentally result in the computational time complexity of the scoring method, nor does it improve its performance in direction discrimination.\par

Generally speaking, there are two kinds of scoring functions, one is the scoring function based on Dirichlet distribution, such as K2 \cite{13} and BDe \cite{14}, and the other is the scoring function based on information theory, such as AIC \cite{40} and LLD. Here we focus on introducing and improving the scoring function based on information theory. In fact, both AIC and BIC impose penalty terms on the basis of LLD, and the LLD form is given in formula \ref{f8}.

\begin{equation}\label{f7}
g_{LLD}(\mathcal{G},D)=\sum_{i=1}^{n}\sum_{j=1}^{q_{i}}\sum_{k=1}^{r_{i}}N_{ijk}log(\frac{N_{ijk}}{N_{ij}})
\end{equation}

Among them, $N_{ijk}$ represents the number of samples in the sample set that satisfy the nodes $X_i=x_i$,$p_a(X_i)=pa_j$ in BN, $N_{ij}=\sum_{k}N_{ijk}$ . Multiplying the left and right sides of formula \ref{f10} by $-\frac{1}{N}$ at the same time, the following changes are obtained:

\begin{equation}
\nonumber
\begin{aligned}
& -\frac{1}{N}g_{LLD}(\mathcal{G},D)=-\sum_{i=1}^{n}\sum_{j=1}^{q_{i}}\sum_{k=1}^{r_{i}}\frac{N_{ijk}}{N}log(\frac{N_{ijk}}{N_{ij}})\\
&\qquad\qquad\qquad =-\sum_{i=1}^{n}\sum_{j=1}^{q_{i}}\sum_{k=1}^{r_{i}}\frac{N_{ijk}}{N}log(\frac{\frac{N_{ijk}}{N}}{\frac{N_{ij}}{N}})\\
&\qquad\qquad\qquad =-\sum_{i=1}^{n}\sum_{j=1}^{q_{i}}\sum_{k=1}^{r_{i}}P(X_i=x_k,p_a(X_i)=pa_j)log(\frac{P(X_i=x_k,p_a(X_i)=pa_j)}{P(p_a(X_i)=pa_j)})
\end{aligned}
\end{equation}

\begin{equation}
\label{f8}
g_{LLD}(\mathcal{G},D)=-N*\sum_{i=1}^{n}H(X_i|p_a(X_i))
\end{equation}

Formula \ref{f8} associates LLD with the sum of conditional mutual information of all nodes under the condition of their parent node-set, obviously, LLD is negatively correlated with the latter. We know that there is a relationship between conditional information entropy and mutual information, that is: $H(X|Y)=H(X)-I(X,Y)$, then formula \ref{f8} can be further transformed into formula \ref{f9}.

\begin{equation}
\label{f9}
\begin{aligned}
& g_{LLD}(\mathcal{G},D)=-N*\sum_{i=1}^{n}(H(X_i)-I(X_i,p_a(X_i)))\\
& \qquad\qquad=-N*\sum_{i=1}^{n}H(X_i)+N*\sum_{i=1}^{n}I(X_i,p_a(X_i))
\end{aligned}
\end{equation}

Under a given data set, the information entropy of each node in the network is a fixed value, but as the determined network structure is different, the parent node-set of each node is also different. This makes the mutual information of each node and its parent node-set not a constant value under the condition of uncertain network structure. According to formula \ref{f9}, it can be seen that the sum of the mutual information of all nodes and their node-sets of parents is an important factor affecting the overall score of a network. This shows that there is a close relationship between the scoring function, conditional information entropy, and mutual information. Using the decomposability of the scoring function to observe the score of a specific node, there are:

\begin{equation}
\label{f10}
g_{LLD}(\mathcal{G},D)=-N*H(X_i|p_a(X_i))
\end{equation}

The meaning of formula \ref{f10} can be analyzed from the perspective of information theory. $H(X_i|p_a(X_i))$ can be understood as, under the condition that the node-set of parents $(p_a(X_i))$ of $X_i$ is known, the remaining unknown information contained in $X_i$ quantity. The smaller the remaining amount of unknown information of $X_i$, the higher the influence of $p_a(X_i)$ as the node-set of parents on $X_i$, that is, the higher the possibility of $p_a(X_i)$ as the node-set of parents of $X_i$. For the sake of convenience, the conditional information entropy is multiplied by the negative constant, and the formula \ref{f10} becomes the larger the value, the better. Then, for a node $X$, when the direction of the edge between Y is not known and it is assumed that there is an edge connection between $X$ and $Y$, the direction between $X$ and $Y$ can be determined by judging the size of $-N*H(X|Y)$ and $-N*H(Y|X)$? The accuracy of this direct comparison method is not high, and any evaluation method must fully consider the unfavorable factors that may affect its results. For example, in the relationship between information entropy and mutual information between nodes shown in Fig.\ref{fig2}, when node $Y$ itself has a larger information entropy, the possibility that its internal information covers more information of node $X$ is also higher. This drives large mutual information between $X$ and $Y$, which makes the value of $-N*H(X|Y)$ too large, and the value of $-N*H(Y|X)$ tends to a relatively small result when the $X$ information entropy is small, which affects the comparison of the two. To eliminate the influence of the parent node's information entropy on the formula \ref{f10}, it is necessary to further impose a penalty term to optimize the evaluation method. The formula \ref{f11} is an optimization of the formula \ref{f10}.\par

For a node $X$, when it is known that $Y$ has a direct relationship with $X$, the two-node (TN) scoring formula for $Y$ as the parent node of $X$ is:

\begin{equation}
\label{f11}
g_{TN}(X,Y)=-N*H(X|Y)-\lambda*\frac{I(X,Y)}{H(X,Y)}*H(Y)
\end{equation}

Where $\lambda$ is the penalty coefficient, which belongs to an unknown hyperparameter, and $\frac{I(X,Y)}{H(X,Y)}$ is the degree of influence on $X$ when $Y$ is the parent-node of $X$. In the experiment, the value of the hyperparameter $\lambda$ fluctuates greatly, and its value is always proportional to the sample N and inversely proportional to the number of value types of the target node $X$. To simplify the value range of hyperparameters, this paper transforms hyperparameters into a semi-self-adjusting hyperparameter, namely:$\lambda=\hat{\lambda}*log(N,D|X|)$, where $D|X|$ is the number of value types of node $X$. The experiments in Section \ref{sec5} show that after semi-self-tuning, the optimal value range of the new hyperparameter $\hat{\lambda}$ is limited to the range (0,1). \par

So far, for two nodes $X$ and $Y$ that are known to have a direct connection, the direction of the edge between the two can be judged by TN score:

\begin{equation}
\nonumber
g_{TN}(X,Y)=-N*(X|Y)-\hat{\lambda}*log(N,D|X|)*\frac{I(X,Y)}{H(X,Y)}*H(Y)
\end{equation}
\begin{equation}
\nonumber
g_{TN}(Y,X)=-N*(Y|X)-\hat{\lambda}*log(N,D|Y|)*\frac{I(X,Y)}{H(X,Y)}*H(X)
\end{equation}

When $g_{TN}(X,Y)>g_{TN}(Y,X)$, it is considered as $Y \to X$; otherwise, there is $Y\gets X$. To test the effectiveness of the new scoring function, we use the SPORTS dataset in literature \cite{37} to make two sets of tests. The two groups of target-nodes are ATshotsOnTarget (abbreviated as AOT) and ATgoals (abbreviated as ATg), and the node-sets of parents and children of these two nodes in the undirected graph are $(RD,ATs,ATg)$ and $(RD,AOT,HDA)$ respectively. The true node-sets of parents of the two target-nodes are $(RD,ATs)$ and $(RD,AOT)$, respectively. In the real data, RD refers to RDlevel, ATs refers to ATshots, and the hyperparameter of TN score is $\hat{\lambda}$ =0.2. The specific results are recorded in table \ref{t2}.

\begin{table}[ht]
\centering
\caption{The orientation discrimination of ATshotsOnTarget is based on the TN score}
\setlength{\tabcolsep}{1mm}
\begin{tabular}{cccccc}
\hline
 $g_{TN}(AOT,RD)$ & $g_{TN}(RD,AOT)$ & $g_{TN}(AOT,ATs)$ & $g_{TN}(ATs,AOT)$ & $g_{TN}(AOT,ATg)$ & $g_{TN}(ATg,AOT)$\\
 \hline
 -1.979 & -2.751 & -1.824 & -1.913 & -1.914 & -1.909\\
 \hline
\end{tabular}
\label{t2}
\end{table}

The node-set of parents of AOT is $(RD,ATs)$, and the node-set of parents of ATg is $(RD,AOT)$, which is consistent with the real results. Compared with other traditional scoring methods, the biggest advantage of TN score is that the operation speed in the direction of the judgment edge is much faster than the traditional scoring method, and the judgment accuracy is not inferior to the traditional score-based method.

\begin{table}[ht]
\centering
\caption{The orientation discrimination of ATgoals is based on the TN score}
\setlength{\tabcolsep}{1mm}
\begin{tabular}{cccccc}
\hline
 $g_{TN}(ATg,RD)$ & $g_{TN}(RD,ATg)$ & $g_{TN}(ATg,AOT)$ & $g_{TN}(AOT,ATg)$ & $g_{TN}(ATg,HDA)$ & $g_{TN}(HDA,ATg)$\\
 \hline
 $-1.971$ & $-2.744$ & $-1.909$ & $-1.914$ & $-1.772$ & $-1.451$\\
 \hline
\end{tabular}
\label{t3}
\end{table}

\section{Framework of MCME algorithm}
\label{sec4}

\,\,\,\,\,\,\,Based on the above theoretical innovation preparation, this chapter will introduce the main ideas and framework structure of the MCME algorithm in detail.

\subsection{Idea of MCME algorithm}
\label{sec4.1}

\,\,\,\,\,\,\,Based on the shortcomings and improvements of the current hybrid algorithms discussed in section \ref{sec3}, we propose the overall idea of the MCME algorithm. The MCME algorithm is a new hybrid algorithm proposed by us. It improves some of the shortcomings of the current hybrid algorithm, but still maintains the overall steps of the hybrid algorithm: network skeleton learning and edge direction discrimination.

\begin{figure}[ht] 
\centering 
\includegraphics[width=0.6\textwidth]{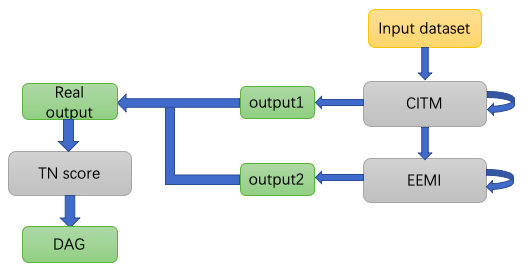}
\caption{Flowchart of the search of the MCME algorithm} 
\label{fig3}
\end{figure}

\subsubsection{Skeleton Learning of MCME Algorithm}
\label{sec4.1.1}

\,\,\,\,\,\,\,According to the improvements mentioned in section \ref{sec3.1} and \ref{sec3.3}, in the skeleton learning of the MCME algorithm, we include both the CI tests with memory and the entropy-eliminated mutual information, two methods of extracting associated nodes. In the process of CI test with memory, the number of memory-erased can be considered as the number of repeated CI tests, which ensures that the associated nodes of the target node are comprehensively searched. This memory can be extended to the process of entropy-eliminated mutual information. When the memory capacity is set greater than 1, the mutual information and information entropy in the entropy-eliminated mutual information will become conditional mutual information and conditional information entropy. The specific process is shown in Fig.\ref{fig3}, and the detailed calculation process is shown in Algorithm \ref{alm3}.

In Fig.\ref{fig3}, with the input of the data set, the CITM (conditional independence tests with memory) layer first obtains an associated node output output1 of each node, and then the remaining nodes and data are input into the EEMI (entropy-eliminated mutual information) layer to get the complementary associated node output output2. The two outputs are merged into the final Real output, which is the learned undirected graph of the network, and then the undirected graph is transformed into a directed graph through TN score, and no ring structure is guaranteed in the process. The specific processes of CITM and EEMI are shown in Fig.\ref{fig4} and Fig.\ref{fig5}.

\subsubsection{CITM layer}
\label{sec4.1.2}
\begin{figure}[ht] 
\centering 
\includegraphics[width=0.6\textwidth]{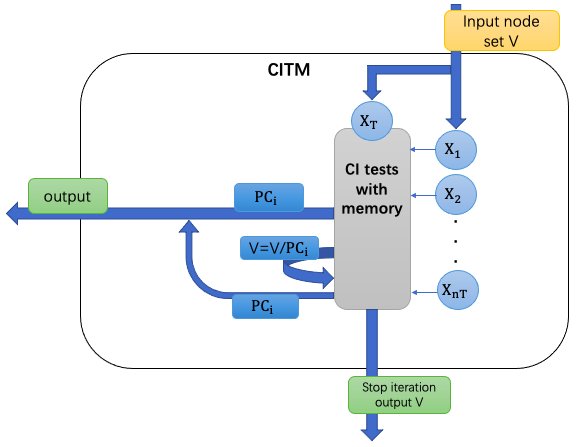}
\caption{Flowchart of CITM} 
\label{fig4}
\end{figure}

\,\,\,\,\,\,\,In Fig.\ref{fig4}, with the input of the target node $X_T$ and the node-set $\textbf{V}$ to be tested, the associated node $X_i$ in $\textbf{V}$ of $X_T$ is screened out through the CI tests with memory, and then the correction operation is performed on $\textbf{V}$, that is, $\textbf{V}=\textbf{V}-X_i$. If the upper memory limit of $X_T$ is not reached, $X_i$ is used as a condition node, and the CI tests with memory are continued on $X_T$ and $\textbf{V}$. Repeat the above process, when the upper memory limit of $X_T$ is reached, save the associated node set of $X_T$ that has been screened out, and clear the memory of $X_T$. If the number of memory-erased is reached, the saved associated node-set will be output, and the remaining node-set $\textbf{V}$ will be output, and enter the EEMI layer as an output. If the number of erasures is not reached, repeat the above process, and collect and save the associated nodes of $X_T$  each time until the number of erasures is reached. The detailed calculation process is shown in Algorithm \ref{alm1}.

\subsubsection{EEMI layer}
\label{sec4.1.3}
\begin{figure}[ht] 
\centering 
\includegraphics[width=0.6\textwidth]{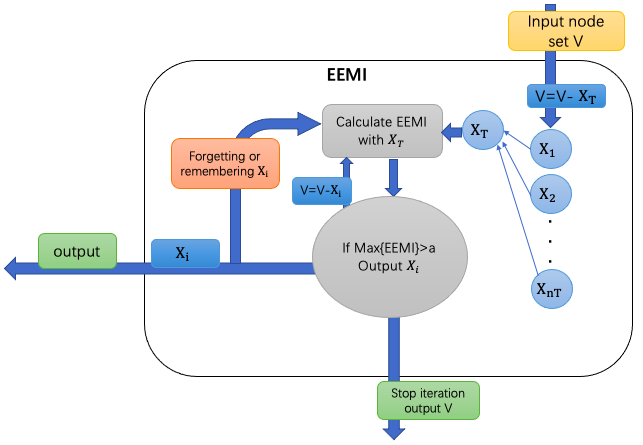}
\caption{Flowchart of EEMI} 
\label{fig5}
\end{figure}

\,\,\,\,\,\,\,If the memory-erased times of the CITM layer are set to 0, the process of filtering the node-set of parents and children of the target node will not go through the CITM layer but directly through the EEMI layer. If the number of memory-erased is greater than 0, then the node-set $\textbf{V}$ flowing into the EEMI layer is the output $\textbf{V}$ from the CITM layer. In the EEMI shown in Fig.\ref{fig5}, each candidate node is paired with the target node $X_T$ to calculate the entropy elimination mutual information, and select the maximum value and the threshold $\alpha$ to compare the size. If it is greater than $\alpha$, the corresponding $X_i$ is selected, and then $\textbf{V}$ is trimmed, that is, $\textbf{V}=\textbf{V}-X_i$. Iterates according to the memory capacity and memory-erased times as described in the CITM layer, until the end of the iteration or when the maximum entropy elimination mutual information does not satisfy the condition greater than $\alpha$, the final node-set of parents and children and the remaining node-set $\textbf{V}$ is output. The detailed calculation process is shown in Algorithm \ref{alm2}.

The memory capacity mentioned in the above CITM layer and EEMI layer is generally less than or equal to 2, and the number of memory-erased is appropriately adjusted according to the selected memory capacity. The two indicators will be reflected in the MCME algorithm as hyperparameters. After performing the associated node search in section \ref{sec4.1.1} for each node in the network, it is necessary to integrate the network skeleton to obtain a complete Bayesian network undirected graph. According to the TN score mentioned in section \ref{sec3.4}, a complete DAG can be obtained by discriminating the direction of the edges of the nodes directly related to each other in the undirected graph.

\subsection{Design details of MCME algorithm}
\label{sec4.2}

\,\,\,\,\,\,\,Algorithm \ref{alm1} faces three parameters, namely, the number of memory-erased $CTlayer$, the memory capacity $CTmemory$ and the significance level $CTalpha$. In general, $CTalpha=0.01$ and $CTmemory$ usually takes 1 or 2. $CTlayer$ needs to be continuously debugged according to specific data.

\begin{algorithm}[ht]
\KwIn{Data $D$, Target node $T$, Candidate node-set $\textbf{V}$, Memory-erased times $CTlayer$, Memory capacity $CTmemory$, Significance level $CTalpha$
}
\KwOut{Node-set of Parents and children of $T$, Node-set of Residual}
time=0;CPC=$\varnothing$ \par
while time $< CTlayer$ or CPC not is changed:\par
\qquad time+=1;cpc=$\varnothing$\par
\qquad while cpc not is changed or $\vert cpc\vert$ $< CTmemory$:\par
\qquad\qquad x=$arg\min_{v\in \textbf{V}}Assoc(T,v|cpc)$\par
\qquad\qquad F=$min_{v\in \textbf{V}}Assoc(T,v|cpc)$\par
\qquad\qquad if F $< CTalpha$:\par
\qquad\qquad\qquad cpc.append(x)\par
\qquad while cpc not is changed:\par
\qquad\qquad x=$arg\max_{v\in cpc}Assoc(T,v|cpc-v$)\par
\qquad\qquad F=$\max_{V\in cpc}Assoc(T,v|cpc-v)$\par
\qquad\qquad if F $> CTalpha$:\par
\qquad\qquad\qquad cpc.remove(x)\par
\qquad $CPC=CPC$ $\cup cpc$\par
\qquad $\textbf{V}=\textbf{V}-cpc$\par
return CPC,$\textbf{V}$\par

\caption{CITM Algorithm}
\label{alm1}
\end{algorithm}

\begin{algorithm}[ht]
\KwIn{Data $D$, Target node $T$, Candidate node-set $\textbf{V}$, Memory-erased times $EElayer$, Memory capacity $EEmemory$, Judgment threshold $EEalpha$
}
\KwOut{Node-set of Parents and children of $T$, Node-set of Residual}
time=0;CPC=$\varnothing$;cpc=$\varnothing$ \par
while time $< EElayer$ or CPC not is changed:\par
\qquad time+=1\par
\qquad if $\vert cpc\vert$ =$EEmemory$:\par
\qquad\qquad cpc=$\varnothing$\par
\qquad x=$arg\max_{v\in \textbf{V}}EEMI(T,v|cpc)$\par
\qquad F=$\max_{v\in \textbf{V}}EEMI(T,v|cpc)$\par
\qquad if F $> EEalpha$:\par
\qquad\qquad cpc.append(x)\par
\qquad\qquad \textbf{V}=\textbf{V}-x\par
\qquad $CPC=CPC\cup cpc$\par
return CPC,\textbf{V}
\caption{EEMI Algorithm}
\label{alm2}
\end{algorithm}

There are also three parameters in Algorithm \ref{alm2}, namely the number of memory-erased $EElayer$, the judgment threshold $EEalpha$ and the memory capacity $EEmemory$. $EEmemory$ usually takes 1 or 0, and $EEalpha$ generally takes 0. At this time, Algorithm \ref{alm2} becomes the candidate node with the largest entropy-eliminated mutual information with the target node every time. The value of $EElayer$ still needs to be judged according to specific data.

\begin{algorithm}[ht]
\KwIn{Data $D$, Target node $T$, Candidate node-set $\textbf{V}$}
\KwOut{Node-set of Parents and children of $T$}
output1,Residualnodeset1=CITM($D$, $T$, $\textbf{V}$, $CTlayer$, $CTmemory$, $CTalpha$)\par
output2,Residualnodeset2=EEMI($D$,$T$, Residualnodeset1,
$EElayer$, $EEmemory$, $EEalpha$)\par
$CPC=output1\cup output2$\par
return CPC\par
\caption{MCME Algorithm}
\label{alm3}
\end{algorithm}

In Algorithm \ref{alm3}, by listing CITM and EEMI as two hierarchical relationships, the nodes in the node set that may be related to the target node are fully extracted. The screening strength of associated nodes can be controlled by adjusting the parameters in CITM and EEMI. The selection rules of parameters for different datasets are introduced in detail in Section \ref{sec5}.

\begin{algorithm}[ht]
\KwIn{Data $D$, All nodes of data $AN$, Penalty coefficient of TN score $\hat{\lambda}$}
\KwOut{Directed acyclic graph of learning $DAG$}
DAG=dict
for $T$ in $AN$:\par
\qquad $\textbf{V}$=$AN-T$\par
\qquad CPC= MCME($D$,$T$,$\textbf{V}$)\par
\qquad $DAG_{T}$=$\varnothing$\par
\qquad for cpc in CPC:\par
\qquad\qquad $g_{TN}(T,cpc|\hat{\lambda})> g_{TN}(cpc,T|\hat{\lambda})$:\par
\qquad\qquad\qquad $DAG_{T}$.append(cpc)\par
\qquad DAG[T]=$DAG_{T}$\par
return DAG\par
\caption{Generate DAG}
\label{alm4}
\end{algorithm}

In Algorithm \ref{alm4}, $\hat{\lambda}$ is the penalty coefficient in the TN score, and its value is generally in the interval (0, 1). The function of Generate DAG is to convert the undirected graph generated by the MCME algorithm into a directed graph, and avoid generating a ring structure during the conversion process. Finally, the directed graph generated by the TN score is the Bayesian network structure learned by the model from the data.\par

\section{Experimental evaluation}
\label{sec5}

\,\,\,\,\,\,\,According to the introduction of the MCME algorithm in Section \ref{sec4}, after a given data, if you want to learn the network structure by applying the MCME algorithm, you first need to determine the values of seven parameters, namely $CT/EEmemory$, $CT/EElayer$, $CT/EEalpha$, and $\hat{\lambda}$. First, the influence of parameter values on network evaluation is determined through experiments, and then the performance of MCME algorithm, MMHC \cite{22} algorithm, and HC \cite{41} algorithm is compared based on six Bayesian networks. All experiments were run on a host with 8GB RAM, dual-core Intel Core i5, and the operating system was Mac. \par

\subsection{Experimental design}
\label{sec5.1}

\,\,\,\,\,\,\,In the process of parameter value experiment, since our fundamental problem is to generate a network that is as close as possible to the original network, two indicators, $tureadd\%$ and $falseadd\%$, are used to evaluate the generated network. $tureadd\%$ represents the percentage of correctly added edges in the generated network to the total number of edges in the original network, and $falseadd\%$ represents the percentage of incorrectly added edges in the generated network to the total number of edges in the original network.\par

In order to explore the influence of parameter values on the generation network, four networks of ASIA (Number of nodes: 8, Number of edges: 8), SPORTS (Number of nodes: 9, Number of edges: 15), PROPERTY (Number of nodes: 27, Number of edges: 31), and ALARM (Number of nodes: 37, Number of edges: 46) are used to test the parameters. In order to test the adaptability of the algorithm to small sample size, the sample size of all experiments is limited to 1000. In the comparative experiment, in addition to the above four networks, the FORMED network (Number of nodes: 88, Number of edges: 138) and the PATHFINDER network (Number of nodes: 109, Number of edges: 195) are additionally added. The above six network structures and datasets are all provided by literature \cite{33}.

\subsection{Parameter value}
\label{sec5.2}

\,\,\,\,\,\,\,The use of CITM or EEMI alone does not meet the needs of the paper, and usually the experiments are a mixture of the two methods. During the experiment, the values of $EEalpha$ and $EEmemory$ are fixed at 0.55 and 1. Here, the values of $EEalpha$ and $EEmemory$ are fixed to control the parameter changes during the experiment and reduce the number of experiments. In practical applications, the size of $EEalpha$ and $EEmemory$ can be adjusted according to the number of edges in the network structure learned by the MCME algorithm. For example, when the number of edges in a network generated by the MCME algorithm is small, the value of $EEalpha$ should be appropriately reduced; otherwise, the value of $EEalpha$ should be increased. When the generated network structure is obviously unrealistic, the value of $EEmemory$ can be increased to increase the accuracy of the learned network structure. In the process of exploring the law of parameter value, the memory capacity ($CTmemory$ takes 1 or 2) and the value of the significance level in the CITM algorithm are changed, and the value of the hyperparameter $\hat{\lambda}$ of the TN scoring function is constantly changed. The experimental results are recorded in table \ref{t3} and table \ref{t4}. Due to the large variety of parameters, the two tables only record the range of parameter values with better results.

\begin{table}[ht]
\centering
\caption{The parameters of MCME algorithm is based on ASIA and SPORTS network(N=1000).}
\setlength{\tabcolsep}{1mm}
\begin{tabular}{cccccccc}
\hline
$CT/EElayer$ & $CT/EEmemory$ & $CT/EEalpha$ & $\hat{\lambda}$ &  \multicolumn{2}{c}{ASIA} & \multicolumn{2}{c}{SPORTS}\\
& & & &  $truedd\%$ & $falseadd\%$ & $trueadd\%$ & $falseadd\%$\\
\hline
1/1 & 1/1 & 0.01/0.55 & 0.1 & 37.5\% & 37.5\% & \textbf{33.3}\% & \textbf{0}\\
1/1 & 1/1 & 0.01/0.55 & 0.2 & 50\% & 25.5\% & 33.3\% & 0\\
1/1 & 1/1 & 0.01/0.55 & 0.3 & \textbf{62.5}\% & \textbf{12.5}\% & 33.3\% & 0\\
1/1 & 1/1 & 0.01/0.55 & 0.4 & 62.5\% & 12.5\% & 33.3\% & 0\\
1/1 & 1/1 & 0.01/0.55 & 0.5 & 50\% & 25\% & 33.3\% & 0\\
1/1 & 1/1 & 0.01/0.55 & 0.6 & 50\% & 25\% & 26.7\% & 6.7\%\\
1/1 & 2/1 & 0.01/0.55 & 0.1 & 35.5\% & 50\% & \textbf{53.3}\% & \textbf{6.7}\%\\
1/1 & 2/1 & 0.01/0.55 & 0.2 & 50\% & 37.5\% & 53.3\% & 6.7\%\\
1/1 & 2/1 & 0.01/0.55 & 0.3 & \textbf{62.5}\% & \textbf{25}\% & 53.3\% & 6.7\%\\
1/1 & 2/1 & 0.01/0.55 & 0.4 & 62.5\% & 25\% & 53.3\% & 6.7\%\\
1/1 & 2/1 & 0.01/0.55 & 0.5 & 50\% & 37.5\% & 53.3\% & 6.7\%\\
1/1 & 2/1 & 0.01/0.55 & 0.6 & 50\% & 37.5\% & 46.7\% & 13.3\%\\
1/1 & 1/1 & 0.001/0.55 & 0.1 & 25\% & 37.5\% & \textbf{33.3}\% & \textbf{0}\\
1/1 & 1/1 & 0.001/0.55 & 0.2 & 37.5\% & 25\% & 33.3\% & 0\\
1/1 & 1/1 & 0.0=1/0.55 & 0.3 & \textbf{50}\% & \textbf{12.5}\% & 33.3\% & 0\\
1/1 & 1/1 & 0.001/0.55 & 0.4 & 50\% & 12.5\% & 33.3\% & 0\\
1/1 & 1/1 & 0.001/0.55 & 0.5 & 37.5\% & 25\% & 33.3\% & 0\\
1/1 & 1/1 & 0.001/0.55 & 0.6 & 37.5\% & 25\% & 26.7\% & 6.7\%\\
1/1 & 2/1 & 0.001/0.55 & 0.1 & 25\% & 37.5\% & \textbf{53.3}\% & \textbf{6.7}\%\\
1/1 & 2/1 & 0.001/0.55 & 0.2 & 37.5\% & 25\% & 53.3\% & 6.7\%\\
1/1 & 2/1 & 0.001/0.55 & 0.3 & \textbf{50}\% & \textbf{12.5}\% & 53.3\% & 6.7\%\\
1/1 & 2/1 & 0.001/0.55 & 0.4 & 50\% & 12.5\% & 53.3\% & 6.7\%\\
1/1 & 2/1 & 0.001/0.55 & 0.5 & 37.5\% & 25\% & 53.3\% & 6.7\%\\
1/1 & 2/1 & 0.001/0.55 & 0.6 & 37.5\% & 25\% & 46.7\% & 13.3\%\\
\hline
\end{tabular}
\label{t4}
\end{table}

\begin{table}[ht]
\centering
\caption{The parameters of the MCME algorithm is based on the PROPERTY and ALARM networks(N=1000).}
\setlength{\tabcolsep}{1mm}
\begin{tabular}{cccccccc}
\hline
$CT/EElayer$ & $CT/EEmemory$ & $CT/EEalpha$ & $\hat{\lambda}$ &  \multicolumn{2}{c}{PROPERTY} & \multicolumn{2}{c}{ALARM}\\
& & & &  $truedd\%$ & $falseadd\%$ & $trueadd\%$ & $falseadd\%$\\
\hline
1/1 & 1/1 & 0.01/0.55 & 0.1 & \textbf{35.5}\% & \textbf{32.2}\% & 8.7\% & 50\%\\
1/1 & 1/1 & 0.01/0.55 & 0.2 & 32.2\% & 35.5\% & 6.5\% & 52.1\%\\
1/1 & 1/1 & 0.01/0.55 & 0.3 & 29\% & 38.7\% & 13\% & 45.7\%\\
1/1 & 1/1 & 0.01/0.55 & 0.4 & 29\% & 38.7\% & 17.4\% & 41.3\%\\
1/1 & 1/1 & 0.01/0.55 & 0.5 & 22.6\% & 45.2\% & 23.9\% & 34.8\%\\
1/1 & 1/1 & 0.01/0.55 & 0.6 & 25.8\% & 41.9\% & 32.6\% & 26.1\%\\
1/1 & 1/1 & 0.01/0.55 & 0.7 & 25.8\% & 41.9\% & \textbf{37}\% & \textbf{21.7}\%\\
1/1 & 1/1 & 0.01/0.55 & 0.8 & 22.6\% & 45.2\% & 37\% & 21.7\%\\
1/1 & 2/1 & 0.01/0.55 & 0.1 & \textbf{45.1}\% & \textbf{35.5}\% & 10.7\% & 80.4\%\\
1/1 & 2/1 & 0.01/0.55 & 0.2 & 41.9\% & 38.7\% & 8.7\% & 82.6\%\\
1/1 & 2/1 & 0.01/0.55 & 0.3 & 38.7\% & 41.9\% & 15.2\% & 76.1\%\\
1/1 & 2/1 & 0.01/0.55 & 0.4 & 38.7\% & 41.9\% & 23.9\% & 67.4\%\\
1/1 & 2/1 & 0.01/0.55 & 0.5 & 32.3\% & 48.4\% & 30.4\% & 60.9\%\\
1/1 & 2/1 & 0.01/0.55 & 0.6 & 35.5\% & 45.2\% & 41.3\% & 50\%\\
1/1 & 2/1 & 0.01/0.55 & 0.7 & 35.5\% & 45.2\% & 45.7\% & 45.7\%\\
1/1 & 2/1 & 0.01/0.55 & 0.8 & 32.2\% & 48.4\% & \textbf{47.8}\% & \textbf{43.5}\%\\
1/1 & 1/1 & 0.001/0.55 & 0.1 & \textbf{35.5}\% & \textbf{22.6}\% & 8.7\% & 45.7\%\\
1/1 & 1/1 & 0.001/0.55 & 0.2 & 32.2\% & 25.8\% & 6.5\% & 47.8\%\\
1/1 & 1/1 & 0.001/0.55 & 0.3 & 29\% & 29\% & 13\% & 45.7\%\\
1/1 & 1/1 & 0.001/0.55 & 0.4 & 29\% & 29\% & 17.4\% & 36.9\%\\
1/1 & 1/1 & 0.001/0.55 & 0.5 & 22.6\% & 25.5\% & 23.9\% & 30.4\%\\
1/1 & 1/1 & 0.001/0.55 & 0.6 & 25.8\% & 32.3\% & 32.6\% & 21.7\%\\
1/1 & 1/1 & 0.001/0.55 & 0.7 & 25.8\% & 32.3\% & \textbf{37}\% & \textbf{17.4}\%\\
1/1 & 1/1 & 0.001/0.55 & 0.8 & 22.6\% & 35.5\% & 37\% & 17.4\%\\
1/1 & 2/1 & 0.01/0.55 & 0.1 & \textbf{45.1}\% & \textbf{25.8}\% & 10.7\% & 69.6\%\\
1/1 & 2/1 & 0.01/0.55 & 0.2 & 41.9\% & 29\% & 8.7\% & 71.7\%\\
1/1 & 2/1 & 0.01/0.55 & 0.3 & 38.7\% & 32.3\% & 15.2\% & 65.2\%\\
1/1 & 2/1 & 0.01/0.55 & 0.4 & 38.7\% & 32.3\% & 23.9\% & 56.5\%\\
1/1 & 2/1 & 0.01/0.55 & 0.5 & 32.3\% & 38.7\% & 30.4\% & 50\%\\
1/1 & 2/1 & 0.01/0.55 & 0.6 & 35.5\% & 35.5\% & 41.3\% & 39.1\%\\
1/1 & 2/1 & 0.01/0.55 & 0.7 & 35.5\% & 35.5\% & 45.7\% & 34.8\%\\
1/1 & 2/1 & 0.01/0.55 & 0.8 & 32.2\% & 38.7\% & \textbf{47.8}\% & \textbf{32.6}\%\\
\hline
\end{tabular}
\label{t5}
\end{table}

\subsection{Experimental comparison}
\label{sec5.3}

\,\,\,\,\,\,\,The focus of this paper is to examine the proximity of the network generated by the algorithm to the original network and the time consumption, so the time consumed by the algorithm to generate the network is marked in the evaluation indicators. Since the size of the global network score is not considered too much, only the BIC score is used to measure different algorithms. To measure the gap between the generation network and the original network more accurately, this paper redefines the Hamming distance ($H(\mathcal{G})$), let $H(\mathcal{G})=A(\mathcal{G})+D(\mathcal{G})$. Among them, $A(\mathcal{G})$ represents the number of edges added by the generated network compared with the original network, and $D(\mathcal{G})$ represents the number of edges added by the generated network less than the original network. For example, define the generated network structure as $\mathcal{G}^1$ and the original network structure as $\mathcal{G}^0$, then there are:
\begin{equation}
\nonumber
\label{f13}
A(\mathcal{G})=\sum_{X_i\in \textbf{V}}\frac{1}{2}\vert p_{a}^{\mathcal{G}^1}(X_i)-p_{a}^{\mathcal{G}^0}(X_i)\vert
\end{equation}

\begin{equation}
\nonumber
\label{f14}
D(\mathcal{G})=\sum_{X_i\in \textbf{V}}\frac{1}{2}\vert p_{a}^{\mathcal{G}^0}(X_i)-p_{a}^{\mathcal{G}^1}(X_i)\vert
\end{equation}

Naturally, the smaller the Hamming distance of a network learned by an algorithm, the closer it is to the original network. Then the MCME algorithm is used to compare with the MMHC and HC algorithms. The MMHC and HC algorithms come from python's pgmpy package \cite{42}, and the MCME algorithm is implemented by python programming software. During the experiment, the internal scoring function of the MMHC algorithm and the HC algorithm adopts the BIC score, and the significance level is controlled at the same level as the $CTalpha$ in the MCME algorithm.\par

Table \ref{t5} shows the comparison results of the three algorithms. It can be clearly seen that the MCME algorithm generates a high-precision network in the face of a small-node network, and the Hamming distance is always optimal. The running time of the MCME algorithm increases steadily with the increase of the number of nodes, and the BIC score level is also comparable to that of the HC algorithm. In the face of a multi-node network, the MCME algorithm gradually shows advantages in terms of operation time and learning accuracy, while the MMHC algorithm takes too long to learn in the face of a network with a large number of nodes, which loses its comparative significance. The MCME algorithm of the first four networks selects the optimal parameters according to the experimental data in table \ref{t3} and \ref{t4}. In table \ref{t5}, the parameters of the FORMED network MCME algorithm are: $CT/EEalpha$=0.001/0.4, $CT/EElayer$=1/1, $\hat{\lambda}$=0.02. PATHFINDER network MCME algorithm parameters are: $CT/EEalpha$=0.001/*, $CT/EElayer$=1/0, $\hat{\lambda}$=0.01.

\begin{table}[ht]
\centering
\caption{The comparison of MCME, HC, and MMHC}
\setlength{\tabcolsep}{2mm}
\begin{tabular}{ccccccc}
\hline
Network & Size of $D$ & BIC of $\mathcal{G}^{0}$ & Evaluation indicators & MCME & HC & MMHC \\
\hline
& & & A($\mathcal{G}^{1}$) & \textbf{1.5} & 2.5 & 2.5\\
& & & D($\mathcal{G}^{1}$) & \textbf{0.5} & 2 & 2 \\
ASIA & 1000 & -3254.92 & H($\mathcal{G}^{1}$) & \textbf{2} & 4.5 & 4.5\\
& & & BIC($\mathcal{G}^{1}$) & -3358.28 & \textbf{-3277.06} & -3534.27\\
& & & time & 2.11s & \textbf{1.28}s & 151.38s\\

& & & A($\mathcal{G}^{1}$) & \textbf{3.5} & 5 & 7.5\\
& & & D($\mathcal{G}^{1}$) & \textbf{0.5} & 2 & 2.5\\
SPORTS & 1000 & -20032.52 & H($\mathcal{G}^{1}$) & \textbf{4} & 7 & 10\\
& & & BIC($\mathcal{G}^{1}$) & -16903.36 & \textbf{-16526.68} & -17696.68\\
& & & time & 66.15s & \textbf{3.49}s & 331.37s\\

& & & A($\mathcal{G}^{1}$) & \textbf{8} & 8.5 & 10.5\\
& & & D($\mathcal{G}^{1}$) & 4 & 8 & \textbf{3}\\
PROPERTY & 1000 & -48553.11 & H($\mathcal{G}^{1}$) & \textbf{12} & 16.5 & 13.5\\
& & & BIC($\mathcal{G}^{1}$) & -45543.77 & \textbf{-40973.96} & -43341.01\\
& & & time & 640.03s & \textbf{47.69}s & 13421.02s\\

& & & A($\mathcal{G}^{1}$) & 14.5 & \textbf{7.5} & /\\
& & & D($\mathcal{G}^{1}$) & 4 & \textbf{8.5} & /\\
ALARM & 1000 & -17508.14 & H($\mathcal{G}^{1}$) & 18.5 & \textbf{16} & /\\
& & & BIC($\mathcal{G}^{1}$) & -20553.44 & \textbf{-17457.82} & /\\
& & & time & \textbf{64.35}s & 86.76s & /\\

& & & A($\mathcal{G}^{1}$) & 53.5 & \textbf{47.5} & /\\
& & & D($\mathcal{G}^{1}$) & \textbf{14.5} & 57.5 & /\\
FORMED & 1000 & -64489.70 & H($\mathcal{G}^{1}$) & \textbf{68} & 105 & /\\
& & & BIC($\mathcal{G}^{1}$) & -77880.64 & \textbf{-64314.05} & /\\
& & & time & \textbf{440.23}s & 1309.69s & /\\

& & & A($\mathcal{G}^{1}$) & \textbf{54} & 96 & /\\
& & & D($\mathcal{G}^{1}$) & \textbf{5} & 31 & /\\
PATHFINDER & 1000 & -327238.53 & H($\mathcal{G}^{1}$) & \textbf{59} & 127 & /\\
& & & BIC($\mathcal{G}^{1}$) & -101019.37 & \textbf{-73914.96} & /\\
& & & time & \textbf{1066.05}s & 13436.12s & /\\
\hline
\end{tabular}
\label{t6}
\end{table}

From table \ref{t5}, we can see that when predicting small network structures (AISA and SPORTS), the prediction accuracy of the MCME algorithm is always the best, but the time consumption is higher than that of the HC algorithm. The MMHC algorithm is lower than the first two algorithms in terms of time consumption and prediction accuracy. When predicting medium-sized network structures (PROPERTY and ALARM), the prediction accuracy of the MCME algorithm is better or comparable to that of the other two algorithms, and as the number of network nodes increases, the time consumption of the MCME algorithm is gradually better than the other two algorithms. When predicting large-scale network structures (FORMED and PATHFINDER), the MCME algorithm begins to show its own advantages. It is far superior to the HC calculation in terms of time consumption and prediction accuracy (the MMHC algorithm is too high in time consumption and loses comparative significance). The overall BIC score of the network structure learned by the MCME algorithm is always lower than that of the HC algorithm, and the HC algorithm is always optimal at the overall network BIC score level. This result is closely related to the use of the BIC score function in the direction discrimination stage of the HC algorithm. The MCME algorithm adopts our proposed TN scoring method to discriminate the direction of edges in the scoring stage, and its overall network BIC score is slightly lower than that of the HC algorithm, which is expected. On the whole, with the increase of the number of network nodes, the time consumption of the MCME algorithm is much lower than that of the HC and MMHC algorithms. And the MCME algorithm is also better than the other two algorithms in the degree of similarity between the learned network structure and the original network structure.

\section{Conclusions and future works}
\label{sec6}
\,\,\,\,\,\,\,In this paper, a novel Bayesian network hybrid learning algorithm MCME is proposed. We improved the traditional CI tests and added memory elements to improve the calculation speed and actuarial accuracy. The related concept of entropy-eliminated mutual information is proposed, and the reliability of entropy-eliminated mutual information is proved by experiments. In the skeleton generation stage, the CI tests with memory elements is combined with the entropy-eliminated mutual information, which improves the reliability of the network skeleton. A new scoring function, two-node scoring, is proposed in the direction discrimination stage.\par

Finally, the experimental comparison is carried out, and the operation time and the accuracy of the learned model are compared with the existing two methods. The performance of the MCME algorithm is always better than or equal to the two algorithms. And the experiment shows that the MCME in the network with a small number of nodes has higher accuracy, but the operation time is relatively slow. With the increase in the number of network nodes, the MCME algorithm shows higher learning accuracy, and the operation time is much shorter than the other two methods.The MCME algorithm can fully meet the needs when learning small network structures. When the number of learned network nodes is large (more than 100 nodes), the MCME algorithm is not enough to learn the complete original network structure, but it is also closer to the original model than the network model learned by other algorithms.\par

In this paper, although the CI tests in the CITM structure adds memory factors, in order to improve the operation speed, each memory-erased is a complete memory-erased, and no part of the memory is retained. In future work, it is possible to try to retain part of the “relatively important” memory (nodes) after each memory-erased. Although it will increase the computational complexity, it is expected to be better than complete memory-erased in the accuracy of searching for associated nodes.In the stage of TN score judgment direction, this paper adopts a semi-parametric self-adjustment mechanism, thereby reducing the value range of hyperparameters. In future work, the self-adjustment of the parameters in the TN score can be realized in combination with the parameter value rules found in table \ref{t4} and \ref{t5}. The last prospect is to continue to optimize the code structure of the MCME algorithm and improve the new performance of the MCME algorithm for the purpose of improving the prediction accuracy of the learned network structure and reducing the time consumption.\par

\section{Funding and/or Conficts of interests/Competing interests}
\label{sec7}
\,\,\,\,\,\,\,We declare that we have no flnancial and personal relationships with other people and organizations, which may improperly affect our work, and that we have no professional or other personal interests of any nature or kind in any products, services and companies.

\bibliographystyle{unsrt}
\bibliography{11}

\begin{thebibliography}{10}

\bibitem{1}
Xiqun Chen, Lingxiao Zhou, and Li~Li.
\newblock Bayesian network for red-light-running prediction at signalized
  intersections.
\newblock {\em Journal of Intelligent Transportation Systems}, 23(2):120--132,
  2019.

\bibitem{2}
Fernando Calle-Alonso, Carlos~J P{\'e}rez, and Eduardo~S Ayra.
\newblock A bayesian-network-based approach to risk analysis in runway
  excursions.
\newblock {\em The Journal of Navigation}, 72(5):1121--1139, 2019.

\bibitem{3}
Timo~JT Koski and John Noble.
\newblock A review of bayesian networks and structure learning.
\newblock {\em Mathematica Applicanda}, 40(1):51--103, 2012.

\bibitem{4}
Marco Scutari, Claudia Vitolo, and Allan Tucker.
\newblock Learning bayesian networks from big data with greedy search:
  computational complexity and efficient implementation.
\newblock {\em Statistics and Computing}, 29(5):1095--1108, 2019.

\bibitem{5}
Anthony~C Constantinou, Yang Liu, Kiattikun Chobtham, Zhigao Guo, and Neville~K
  Kitson.
\newblock Large-scale empirical validation of bayesian network structure
  learning algorithms with noisy data.
\newblock {\em International Journal of Approximate Reasoning}, 131:151--188,
  2021.

\bibitem{6}
Ruihong Xu, Sihang Liu, Qingwang Zhang, Zemeng Yang, and Jianxiao Liu.
\newblock Pewobs: An efficient bayesian network learning approach based on
  permutation and extensible ordering-based search.
\newblock {\em Future Generation Computer Systems}, 128:505--520, 2022.

\bibitem{7}
Alejandro Edera, Yanela Strappa, and Facundo Bromberg.
\newblock The grow-shrink strategy for learning markov network structures
  constrained by context-specific independences.
\newblock In {\em Ibero-American Conference on Artificial Intelligence}, pages
  283--294. Springer, 2014.

\bibitem{8}
Jie Cheng, Russell Greiner, Jonathan Kelly, David Bell, and Weiru Liu.
\newblock Learning bayesian networks from data: An information-theory based
  approach.
\newblock {\em Artificial intelligence}, 137(1-2):43--90, 2002.

\bibitem{9}
Peter Spirtes, Clark Glymour, and Richard Scheines.
\newblock Discovery algorithms for causally sufficient structures.
\newblock In {\em Causation, prediction, and search}, pages 103--162. Springer,
  1993.

\bibitem{10}
Ioannis Tsamardinos, Constantin~F Aliferis, Alexander~R Statnikov, and
  Er~Statnikov.
\newblock Algorithms for large scale markov blanket discovery.
\newblock In {\em FLAIRS conference}, volume~2, pages 376--380. St. Augustine,
  FL, 2003.

\bibitem{11}
George Rebane and Judea Pearl.
\newblock The recovery of causal poly-trees from statistical data.
\newblock {\em arXiv preprint arXiv:1304.2736}, 2013.

\bibitem{12}
Robert Castelo and Alberto Roverato.
\newblock A robust procedure for gaussian graphical model search from
  microarray data with p larger than n.
\newblock {\em Journal of Machine Learning Research}, 7(12), 2006.

\bibitem{13}
Gregory~F Cooper and Edward Herskovits.
\newblock A bayesian method for the induction of probabilistic networks from
  data.
\newblock {\em Machine learning}, 9(4):309--347, 1992.

\bibitem{14}
David Heckerman, Dan Geiger, and David~M Chickering.
\newblock Learning bayesian networks: The combination of knowledge and
  statistical data.
\newblock {\em Machine learning}, 20(3):197--243, 1995.

\bibitem{15}
Gideon Schwarz.
\newblock Estimating the dimension of a model.
\newblock {\em The annals of statistics}, pages 461--464, 1978.

\bibitem{16}
Luis~M De~Campos and Nir Friedman.
\newblock A scoring function for learning bayesian networks based on mutual
  information and conditional independence tests.
\newblock {\em Journal of Machine Learning Research}, 7(10), 2006.

\bibitem{17}
Cassio~P De~Campos and Qiang Ji.
\newblock Efficient structure learning of bayesian networks using constraints.
\newblock {\em The Journal of Machine Learning Research}, 12:663--689, 2011.

\bibitem{18}
Marc Teyssier and Daphne Koller.
\newblock Ordering-based search: A simple and effective algorithm for learning
  bayesian networks.
\newblock {\em arXiv preprint arXiv:1207.1429}, 2012.

\bibitem{19}
Colin Lee and Peter~van Beek.
\newblock Metaheuristics for score-and-search bayesian network structure
  learning.
\newblock In {\em Canadian Conference on Artificial Intelligence}, pages
  129--141. Springer, 2017.

\bibitem{20}
Mauro Scanagatta, Giorgio Corani, and Marco Zaffalon.
\newblock Improved local search in bayesian networks structure learning.
\newblock In {\em Advanced Methodologies for Bayesian Networks}, pages 45--56.
  PMLR, 2017.

\bibitem{21}
Ioannis Tsamardinos, Laura~E Brown, and Constantin~F Aliferis.
\newblock The max-min hill-climbing bayesian network structure learning
  algorithm.
\newblock {\em Machine learning}, 65(1):31--78, 2006.

\bibitem{22}
Ioannis Tsamardinos, Laura~E Brown, and Constantin~F Aliferis.
\newblock The max-min hill-climbing bayesian network structure learning
  algorithm.
\newblock {\em Machine learning}, 65(1):31--78, 2006.

\bibitem{23}
Moninder Singh and Marco Valtorta.
\newblock Construction of bayesian network structures from data: a brief survey
  and an efficient algorithm.
\newblock {\em International journal of approximate reasoning}, 12(2):111--131,
  1995.

\bibitem{24}
Peter Spirtes, Clark Glymour, and Richard Scheines.
\newblock Discovery algorithms for causally sufficient structures.
\newblock In {\em Causation, prediction, and search}, pages 103--162. Springer,
  1993.

\bibitem{25}
Denver Dash and Marek~J Druzdzel.
\newblock A hybrid anytime algorithm for the constructiion of causal models
  from sparse data.
\newblock {\em arXiv preprint arXiv:1301.6689}, 2013.

\bibitem{26}
Luis~M De~Campos, Juan~M Fern{\'a}ndez-Luna, and J~Miguel Puerta.
\newblock An iterated local search algorithm for learning bayesian networks
  with restarts based on conditional independence tests.
\newblock {\em International Journal of Intelligent Systems}, 18(2):221--235,
  2003.

\bibitem{27}
Steffen~L Lauritzen, A~Philip Dawid, Birgitte~N Larsen, and H-G Leimer.
\newblock Independence properties of directed markov fields.
\newblock {\em Networks}, 20(5):491--505, 1990.

\bibitem{28}
Hui Liu, Shuigeng Zhou, Wai Lam, and Jihong Guan.
\newblock A new hybrid method for learning bayesian networks: Separation and
  reunion.
\newblock {\em Knowledge-Based Systems}, 121:185--197, 2017.

\bibitem{29}
Judea Pearl.
\newblock {\em Probabilistic reasoning in intelligent systems: networks of
  plausible inference}.
\newblock Morgan kaufmann, 1988.

\bibitem{30}
Thomas~S Verma and Judea Pearl.
\newblock Equivalence and synthesis of causal models.
\newblock In {\em Probabilistic and Causal Inference: The Works of Judea
  Pearl}, pages 221--236. 2022.

\bibitem{31}
Du-Yih Tsai, Yongbum Lee, and Eri Matsuyama.
\newblock Information entropy measure for evaluation of image quality.
\newblock {\em Journal of digital imaging}, 21(3):338--347, 2008.

\bibitem{32}
B~Gierlichs, L~Batina, P~Tuyls, and B~Preneel.
\newblock Mutual information analysis. a generic side-channel distinguisher.
  ches’08, lncs, 2008.

\bibitem{33}
S~Kullback.
\newblock Information theory and statistics—dover publi.
\newblock {\em Inc., NY}, 1968.

\bibitem{34}
Pater Spirtes, Clark Glymour, Richard Scheines, Stuart Kauffman, Valerio
  Aimale, and Frank Wimberly.
\newblock Constructing bayesian network models of gene expression networks from
  microarray data.
\newblock 2000.

\bibitem{35}
Pater Spirtes, Clark Glymour, Richard Scheines, Stuart Kauffman, Valerio
  Aimale, and Frank Wimberly.
\newblock Constructing bayesian network models of gene expression networks from
  microarray data.
\newblock 2000.

\bibitem{36}
Jerome~L Neapolitan.
\newblock Explaining variation in crime victimization across nations and within
  nations.
\newblock {\em International Criminal Justice Review}, 13(1):76--89, 2003.

\bibitem{37}
Anthony~C Constantinou, Yang Liu, Kiattikun Chobtham, Zhigao Guo, and Neville~K
  Kitson.
\newblock The bayesys data and bayesian network repository.
\newblock {\em Queen Mary University of London: London, UK}, 2020.

\bibitem{38}
Leon Brillouin.
\newblock {\em Science and information theory}.
\newblock Courier Corporation, 2013.

\bibitem{39}
Abbas El~Gamal and Young-Han Kim.
\newblock {\em Network information theory}.
\newblock Cambridge university press, 2011.

\bibitem{40}
Hirotugu Akaike.
\newblock A new look at the statistical model identification.
\newblock {\em IEEE transactions on automatic control}, 19(6):716--723, 1974.

\bibitem{41}
Jos{\'e}~A G{\'a}mez, Juan~L Mateo, and Jos{\'e}~M Puerta.
\newblock Learning bayesian networks by hill climbing: efficient methods based
  on progressive restriction of the neighborhood.
\newblock {\em Data Mining and Knowledge Discovery}, 22(1):106--148, 2011.

\bibitem{42}
Ankur Ankan and Abinash Panda.
\newblock pgmpy: Probabilistic graphical models using python.
\newblock In {\em Proceedings of the 14th python in science conference (scipy
  2015)}, volume~10. Citeseer, 2015.

\end{thebibliography}
\end{document}